%% file: main.tex
\documentclass[acmtog,  screen, review=false ,nonacm]{acmart}

\usepackage{subfig}
\usepackage{multirow}

\AtBeginDocument{%
  \providecommand\BibTeX{{%
    \normalfont B\kern-0.5em{\scshape i\kern-0.25em b}\kern-0.8em\TeX}}}

\setcopyright{acmlicensed}
\copyrightyear{2022}
\acmYear{2022}

\acmJournal{TOG}

\author{Jiayang Bai, Jie Guo\textsuperscript{$\dagger$}, Chenchen Wan, Zhenyu Chen, Zhen He, Shan Yang, Piaopiao Yu, Yan Zhang and Yanwen Guo\textsuperscript{$\dagger$}
}
\thanks{\textsuperscript{$\dagger$} is the corresponding author.}
\affiliation{
\institution{State Key Lab for Novel Software Technology, Nanjing University}
\city{Nanjing}
\country{China}
}
\begin{document}

\title{Deep Graph Learning for Spatially-Varying Indoor Lighting Prediction}

\begin{abstract}
Lighting prediction from a single image is becoming increasingly important in many vision and augmented reality (AR) applications in which shading and shadow consistency between virtual and real objects should be guaranteed. However, this is a notoriously ill-posed problem, especially for indoor scenarios, because of the complexity of indoor luminaires and the limited information involved in 2D images. In this paper, we propose a graph learning-based framework for indoor lighting estimation. At its core is a new lighting model (dubbed DSGLight) based on depth-augmented Spherical Gaussians (SG) and a Graph Convolutional Network (GCN) that infers the new lighting representation from a single LDR image of limited field-of-view. Our lighting model builds 128 evenly distributed SGs over the indoor panorama, where each SG encoding the lighting and the depth around that node. The proposed GCN then learns the mapping from the input image to DSGLight. Compared with existing lighting models, our DSGLight encodes both direct lighting and indirect environmental lighting more faithfully and compactly. It also makes network training and inference more stable. The estimated depth distribution enables temporally stable shading and shadows under spatially-varying lighting. Through thorough experiments, we show that our method obviously outperforms existing methods both qualitatively and quantitatively.

\end{abstract}



\begin{teaserfigure}
  \includegraphics[width=\textwidth]{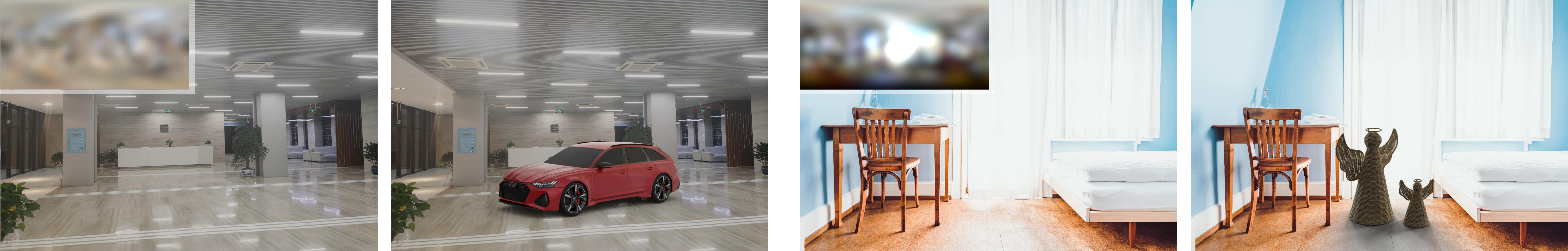}
  \caption{Our method takes a single image as the input and predicts the full and spatially-varying indoor illumination that can be used to generate consistent shading and realistic shadows after inserting virtual objects into the image.}
  \label{fig:insertobjecet}
\end{teaserfigure}

\maketitle

\input{section/ch1}

\input{section/ch2}

\input{section/ch3}

\input{section/ch4}

\input{section/ch5}

\input{section/ch6}

\bibliographystyle{ACM-Reference-Format}
\bibliography{egbib}

\end{document}

%% file: section/ch1.tex
\section{Introduction}

Predicting indoor illumination from a single image is a challenging but meaningful task \cite{https://doi.org/10.1111/cgf.14283}. It is a fundamental step for many applications, ranging from indoor scene understanding to augmented reality (AR). However, this problem is highly ill-posed due to the following reasons. First, indoor illumination stems from a wide range of different luminaires including spot lights, large area lights and indirect lighting. This is quite different from outdoor lighting which is dominated by the sun~\cite{6994881,8099738,8953336,8954446,Yu_2021_ICCV}. Second, a single image captured by an ordinary camera has limited field-of-view (FoV) that may fail to provide sufficient cues of indoor light sources. In fact, many indoor light sources have a local impact and yield spatially-varying lighting effects. This makes them hard to be identified from a single FoV-limited image. Moreover, strong lighting-reflectance ambiguities exist in indoor scenes since different combinations of lighting and surface reflectance may produce the same pixel intensities.

Currently, this problem is addressed in two ways. One line of work assumes the availability of some prior knowledge about the scene, e.g., scene geometries~\cite{1,DBLP:conf/iccv/MaierKCKN17,weber_3dv_18} obtained by depth sensors or multi-view inputs. Prior knowledge is effective for overcoming the ill-posedness of the problem but will limit the applicability of the methods.
A recent trend is to leverage deep learning based solutions accompanied with large training datasets to infer lighting from a single, casually captured image~\cite{DBLP:conf/iccv/GardnerHSGL19,DBLP:journals/corr/GardnerSYSGGL17,8954392,DBLP:conf/cvpr/SongF19,DBLP:conf/cvpr/SrinivasanMTBTS20,zhan2021emlight,zhan2021Needlets}. The lighting models used in these solutions can be divided into two categories. In the first category, the indoor lighting of surrounding environment is stored in a densely sampled environment map~\cite{DBLP:journals/corr/GardnerSYSGGL17}, i.e., image-based lighting (IBL). Unfortunately, inferring a full environment map directly is very difficult due to the high dimensionality of the representation. Alternatively, the second category leverages analytical models with a small number of parameters to fit real-world illumination. For instance, indoor illumination can be approximated by a set of Spherical Harmonics (SH)~\cite{https://doi.org/10.1111/cgf.13561,8954392,Green2003SphericalHL}  or Spherical Gaussians (SG)~\cite{DBLP:conf/iccv/GardnerHSGL19,DBLP:journals/tog/TsaiS06,DBLP:journals/tog/WangRGSG09,sg2add,zhan2021emlight}. Generally, SH works well for low-frequency lighting, but is plagued with ringing artifacts as the SH order increases. Gardner et al.~\shortcite{DBLP:conf/iccv/GardnerHSGL19} proposed using 2-5 SGs to model high-frequency light sources in an indoor scene. Although it outperforms SH-based methods, regressing the positions of floating SGs is unstable for deep learning. Consequently, this method achieves degraded performance as the number of SGs increases. 

To retain the advantages of SGs in compactly encoding high-frequency lighting and enable stable training, we introduce \emph{DSGLight}, a novel SG-based lighting model augmented with depth values. In this model, $N$ SGs are assumed to have fixed positions and are evenly distributed over a unit sphere. We also fix the bandwidth of each SG such that only its color and depth values vary. In this case, predicting indoor lighting boils down to estimating $N$ RGBD values, which is more stable and robust than regressing several free SGs on the sphere. The augmented depth values allow us to generate spatially-varying shading and shadows with respect to the insert location.

Considering the non-Euclidean nature of DSGLight, we design a graph convolution network (GCN) for feature extraction and lighting prediction. GCNs have gained impressive successes in many vision tasks, but are less exploited in lighting prediction. In our network, we first extract features from an input image with vanilla convolutional layers. Then, these features are connected to GCN features with several fully-connected layers. After information exchanging across neighboring nodes, we finally obtain the color and depth for each SG. We demonstrate through extensive experiments that our method outperforms existing methods both qualitatively and quantitatively. The state-of-the-art performance on indoor lighting prediction allows us to achieve more realistic shading results when inserting virtual objects into real indoor scenes (see Fig.~\ref{fig:insertobjecet}).

In summary, our main contributions in this paper are:
\begin{itemize}
    \item A depth-augmented indoor lighting model, i.e., DSGLight, to compactly and faithfully encode spatially-varying lighting in an indoor scene. 
    \item A dedicated graph convolutional network for predicting DSGLight from a single image.
    \item A new dataset with supervision of DSGLight parameters that is constructed to train our network. 
\end{itemize}

%% file: section/ch2.tex
\section{Related Work}

\subsection{Indoor lighting prediction}
Inferring the illumination from 2D images is a classic problem that has been extensively studied in the past decades. Early progress on illumination estimation relied on prior knowledge of scene geometry. For instance, Barron and Malik~\shortcite{1} adopted depth sensors to capture geometric information, Wu et al.~\shortcite{5995388} reconstructed geometry with multi-view stereo, and Larsch et al.~\shortcite{10.1145/2024156.2024191} required users to annotate the geometry. 
Some other works manage to predict illumination from purely 2D image(s). Khan et al.~\shortcite{10.1145/1179352.1141937} presented a method to flip the given image to approximate the rest of the whole panorama, which are out of view. Unfortunately, this only works well in some special cases. 

While deep learning is significantly successful in many graphical and vision tasks, some learning-based methods are proposed to estimate illumination. Hold-Geoffroy et al.~\shortcite{inproceedings14} presented an end-to-end approach that leveraged CNN to predict outdoor illumination represented by a low-dimensional analytical model, significantly outperforming previous traditional methods. Gardner~\shortcite{DBLP:journals/corr/GardnerSYSGGL17} also adopted an end-to-end CNN to regress a non-parametric environment map to represent accurate illumination. Song et al. \shortcite{DBLP:conf/cvpr/SongF19} proposed ``Neural Illumination'' that can predict indoor lighting at a specific locale. Another recent work~\cite{DBLP:conf/cvpr/SrinivasanMTBTS20} leveraged a multi-scale volumetric lighting representation to estimate spatially-coherent illumination from images. 

Close to our work, several methods also leverage multiple SGs to represent indoor lighting \cite{DBLP:conf/iccv/GardnerHSGL19,zhan2021emlight}. However, these methods only capture dominated lights and do not account for spatially-varying environmental lights, resulting in fewer details for rendering specular objects and inconsistent shadows. Moreover, these methods usually leverage CNNs to establish the mapping between input images and SGs. However, CNNs, with a limited receptive field, fail to explore the long-range correlations between SGs.

\subsection{Graph convolution networks}
To deal with non-Euclidean data like graphs, Graph convolution networks (GCNs) provide well-suited solutions which can be divided into two types: spatial-based ~\cite{GCNarticle,4773279,DBLP:journals/corr/NiepertAK16} and spectral-based~\cite{10.5555/3157382.3157527, DBLP:journals/corr/HamiltonYL17, DBLP:conf/iclr/KipfW17}.
Spatial-based methods used in many works are mainly based on the spatial relations of vertices in graphs and they apply spectral convolutions on graph using ideas from graph signal processing~\cite{spetral26}.
As transforming the signal to spectral domains can be expensive, many recent works~\cite{10.5555/3157382.3157527, DBLP:journals/corr/KipfW16} approximate the spectral convolutions using Chebyshev polynomials. Similarly, our network also uses Chebyshev polynomials for spectral convolutions in GCN layers. We use GCNs to extract features from indoor illumination encoded in a non-Euclidean form and establish the connection between 2D images and a SG-based lighting model.

%% file: section/ch3.tex
\section{DSGLight}
\begin{figure}[tb]
  \centering
    \subfloat{\includegraphics[width=0.3\linewidth]{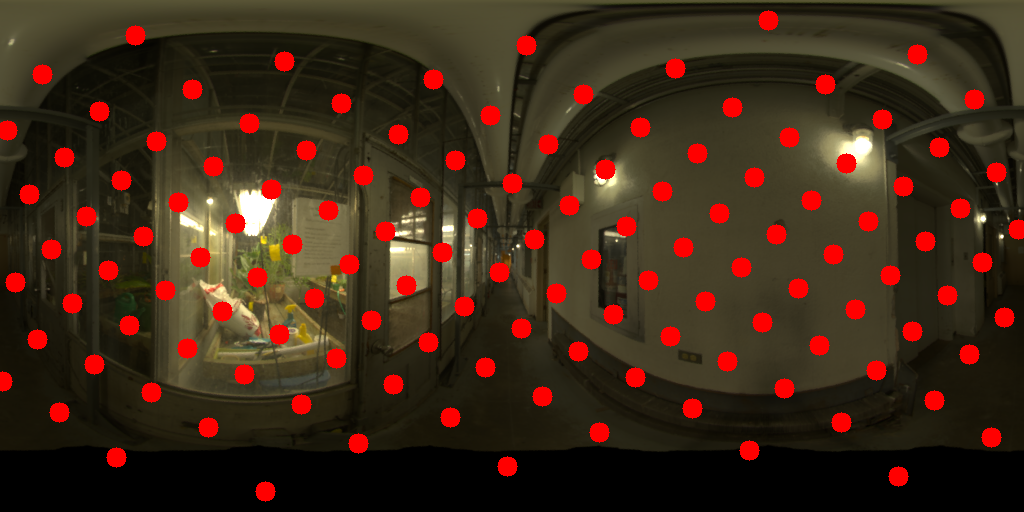}
    } 
    \subfloat{\includegraphics[width=0.3\linewidth]{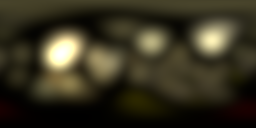}
    } 
    \subfloat{\includegraphics[width=0.3\linewidth]{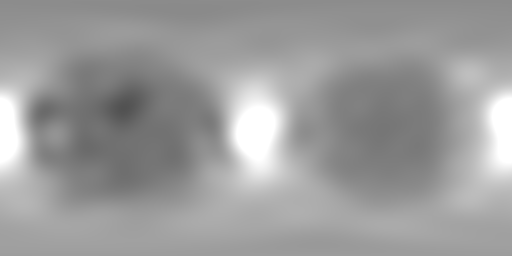}
    }
  \caption{Illustration of DSGLight. Left: centering directions of 128 evenly distributed SGs. Middle: fitted DSGLight (RGB channels) representation for an indoor panorama. Right: fitted DSGLight of depth representation.} \label{fig:DSGLight}
  \vspace{-1.5em}
\end{figure}
\begin{figure*}[phtb]
\centering
\subfloat[Original]{
    \begin{minipage}[t]{0.19\linewidth}
        \centering
        \includegraphics[width=1.0\linewidth]{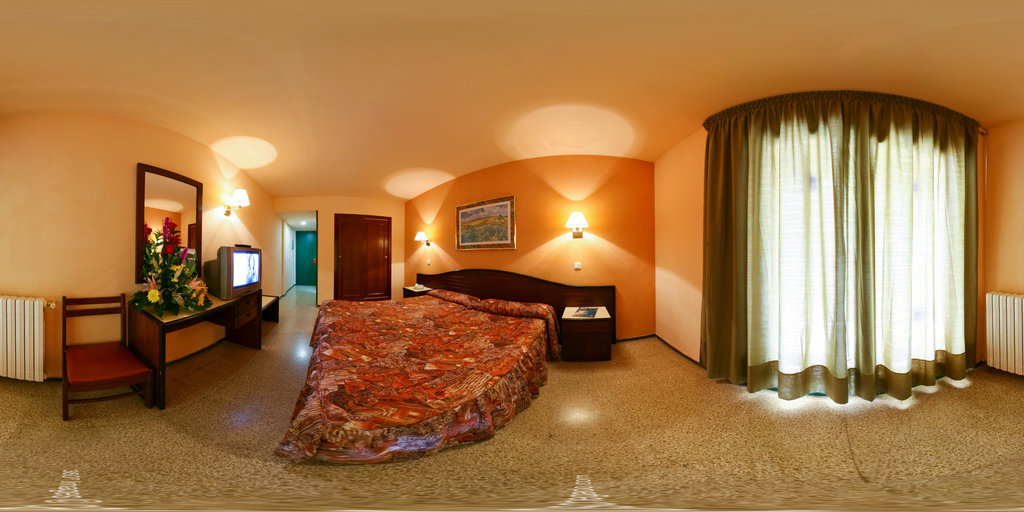}\\
        \vspace{0.02cm}
    \end{minipage}%
}%
\subfloat[DSGLight (RGB channels)]{
    \begin{minipage}[t]{0.19\linewidth}
        \centering
        \includegraphics[width=1.0\linewidth]{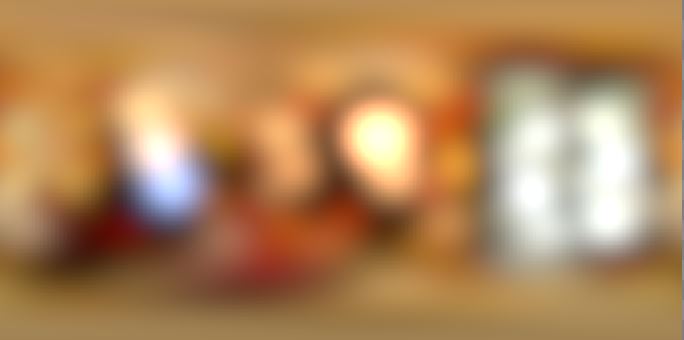}\\
        \vspace{0.02cm}
    \end{minipage}%
}%
\subfloat[~\cite{DBLP:conf/iccv/GardnerHSGL19} (3 SGs)]{
    \begin{minipage}[t]{0.19\linewidth}
        \centering
        \includegraphics[width=1.0\linewidth]{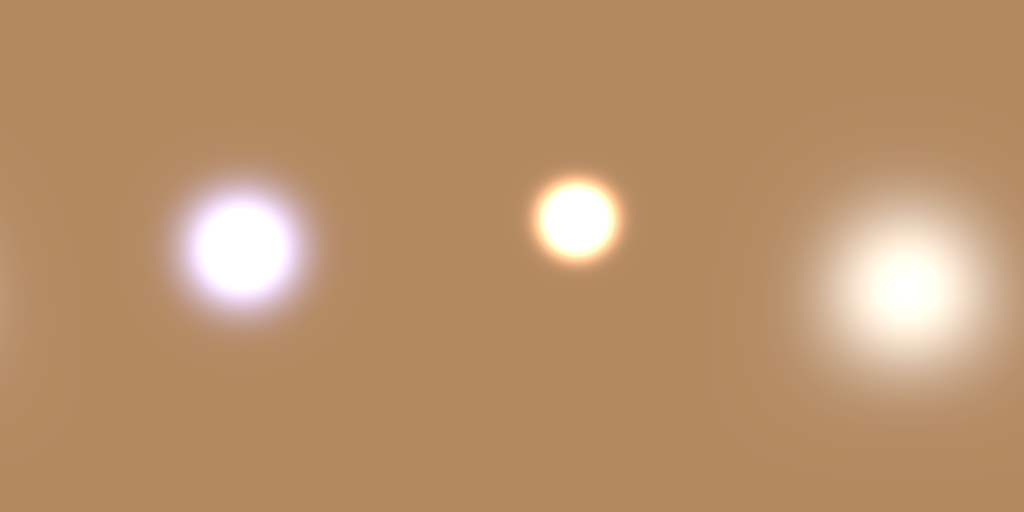}\\
        \vspace{0.02cm}
    \end{minipage}%
}%
\subfloat[~\cite{DBLP:conf/iccv/GardnerHSGL19} (5 SGs)]{
    \begin{minipage}[t]{0.19\linewidth}
        \centering
        \includegraphics[width=1.0\linewidth]{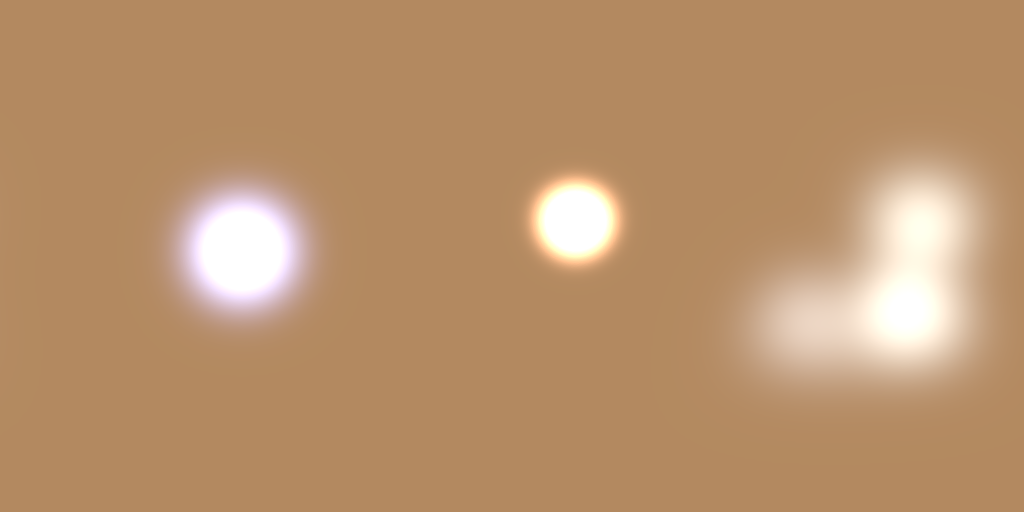}\\
        \vspace{0.02cm}
    \end{minipage}%
}%
\subfloat[~\cite{8954392} (SH) ]{
    \begin{minipage}[t]{0.19\linewidth}
        \centering
        \includegraphics[width=1.0\linewidth]{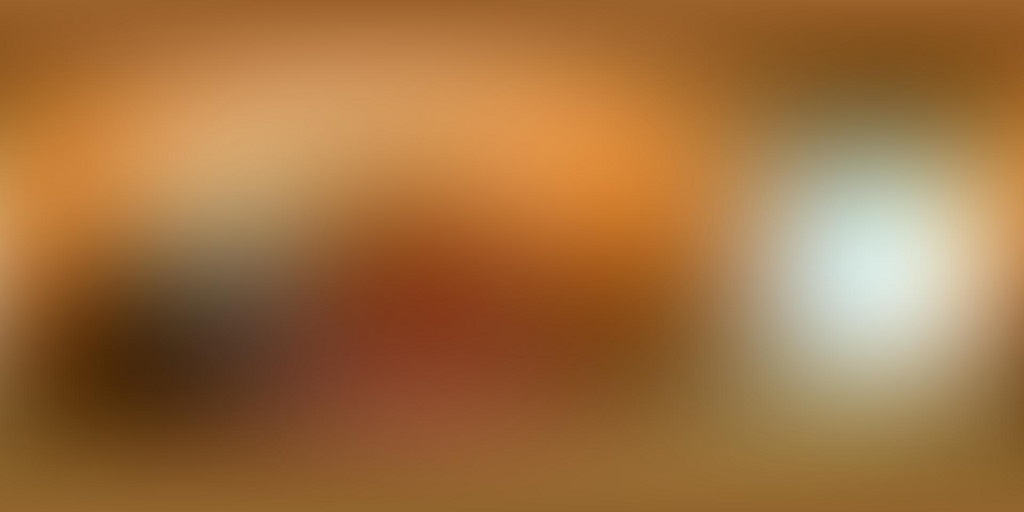}\\
        \vspace{0.02cm}
    \end{minipage}%
}%
\centering
\caption{\label{fig:comparision}Visual comparisons of different lighting representations for an indoor scene.}
\end{figure*}
\subsection{Representation}
DSGLight is a depth-augmented parametric lighting model that leverages a set of discrete SG functions to represent the spatially-varying indoor illumination, as illustrated in Fig.~\ref{fig:DSGLight}. 
We denote DSGLight by $\mathcal{M}=\{m_i |i=1, \dots, N\}$ which contains $N$ vertices.
Each vertex in DSGLight is associated with a spherical Gaussian~\cite{DBLP:journals/tog/WangRGSG09} defined as follows:
\begin{equation}
G( \mathbf{v} ;\mathbf{a}, \lambda, \mu)=\mathbf{a} e^{\lambda (\mathbf{v} \mu - 1)}  \label{SG}
\end{equation}
where $\mu$ is the lobe axis (the centering direction), $\lambda \in  (0,+\infty)$ is the lobe sharpness (the bandwidth), and $\mathbf{a}$ is the lobe amplitude ($\mathbf{a} \in \mathbb{R}^4$ for RGBD information). The direction $\mathbf{v}$ denotes the spherical parameter of the resulting function. Given $m_i=\{\mathbf{a}_i, \lambda_i ,\mu_i \}$ for $i^{th}$ vertex, the overall indoor illumination encoded in a given panorama is compressed into an arbitrary spherical function $f(\mathbf{v})$ as follows:
\begin{equation}
f(\mathbf{v})=\sum _{i}^{N}G(\mathbf{v}; m_i) \label{hdr represent}.
\end{equation}
The corresponding depth panorama is treated in a similar way.
Currently, we set $N$ to 128.

To further reduce the number of parameters, we fix the position $\mu$ and the bandwidth $\lambda$ of each SG. Specifically, 128 nodes are placed evenly over a unit sphere, and they share the same bandwidth. This allows $\mu$ and $v$ to be preset for our model. The distribution of these nodes is shown in Fig.~\ref{fig:DSGLight}. Besides being a more compact representation, such a design stabilizes network training and inferences. As pointed out in~\cite{DBLP:conf/iccv/GardnerHSGL19}, regressing several floating SGs with floating positions and unconstrained bandwidth is unstable, especially for many SGs. As a result, performance decreases from 3 to 5 SGs. With fixed position and bandwidth, inferring DSGLight boils down to predicting $N$ RGBD values, which is easier and more stable than inferring floating SGs.

To make $N$ SGs cover the whole panorama as much as possible and avoid overlapping, we set the bandwidth parameter $\lambda$ as follows:
\begin{equation}
\lambda=\frac{\ln{0.6}}{\cos(\arctan(2/\sqrt{N}))-1}.  \label{alpha}
\end{equation}
Details are provided in the supplemental material.
\subsection{Comparison and analysis}

Previous works have proposed many parametric methods to reduce the number of parameters in representing panoramic lighting, such as SH, SG and IBL. Here, we compare these representations with our DSGLight. One example comparing these representations is provided in Fig.~\ref{fig:comparision}.

Among these methods, IBL is the most accurate. However, regressing the high-dimensional IBL directly may lead to local minimal and it is difficult to converge during training. 
With much fewer parameters, SH is a low-dimensional lighting representation and can be easily predicted with a compact decoder. Unfortunately, SH is too sensitive to its parameters’ variation. Moreover, the network has a tendency to overfit to the ringing artifacts when employing high-order SH functions to preserve high-frequency information.

SG is able to represent high-frequency lighting and avoid ringing artifacts. However, Gardner et al.~\shortcite{DBLP:conf/iccv/GardnerHSGL19} show that it is hard to optimize the positions of SGs, especially when the number of SGs is large. In comparison, DSGLight represents an indoor panorama with a certain number of SG functions which have fixed positions and bandwidth. It is more stable and effective than predicting arbitrary and floating light sources. In addition, as the vertices are spread evenly on the unit sphere and cover the entire sphere, DSGLight can capture both dominated light sources and detailed environmental lighting. This allows high-quality AR rendering with consistent shading and shadows, as well as fine details on specular objects.

\subsection{Dataset preparation}\label{sec:dataset}
To generate the dataset for training, we have to prepare a large number of images and their corresponding DSGLight parameters. Given a HDR panorama, generating ground truth DSGLight can be solved as an optimization problem, i.e., minimizing the difference between illumination/depth constructed by DSGLight and the ground truth panorama: 
\begin{equation}\label{equ:optimization}
\begin{aligned}
\boldsymbol{A}^*=\underset{\boldsymbol{A}}{\arg \min } \sum_{p\in \mathcal{P}}\left(p-\sum_{i=1}^{N} G\left(F(p) ; \mathbf{a}_{i}, \lambda_{i}, \mu_{i}\right)\right)^{2}
\end{aligned}
\end{equation}
where $\boldsymbol{A}=\{\mathbf{a}_{1},\mathbf{a}_{2},\dots, \mathbf{a}_{N}\}$ is the set of lobe amplitudes of SGs, $p$ is a pixel in the ground truth panoramas $\mathcal{P}$, and $F$ maps $p$ to a direction $\mathbf{v}$ in the sphere.
When fixing $\mathcal{P}$, $\mu$ and $\lambda$, the problem boils down to a linear optimization problem, which can be solved easily by nonnegative least square algorithms. 

We currently perform the optimization on two datasets:
\begin{itemize}
    \item \textbf{Laval dataset}~\cite{DBLP:journals/tog/GardnerSYSGGL17} contains 2,200 high resolution indoor panoramas which are fully HDR without any saturation. This dataset also contains depth values for each panorama.
    \item \textbf{SUN360 Dataset}~\cite{DBLP:conf/cvpr/XiaoEOT12} contains 12,000 indoor LDR panoramas captured in a wide range of indoor scenes. For this dataset, we first convert each LDR panorama into a HDR panorama with the trained network of ~\cite{DBLP:journals/tog/EilertsenKDMU17}. This dataset does not contain depth values. Therefore, we can only handle DSGLight with RGB channels when using examples from this dataset. 
\end{itemize}

We further employ Mitsuba renderer~\cite{Mitsuba} to sample several Fov-limited images from each panorama at random elevation between -20\textdegree  and 20\textdegree, azimuth between -180\textdegree  and 180\textdegree and FoV between 60\textdegree  and 80\textdegree. Meanwhile, the corresponding panorama is captured by aligning its center with the view direction.

%% file: section/ch4.tex
\begin{figure*}[htb]
\centering
\includegraphics[scale=0.35]{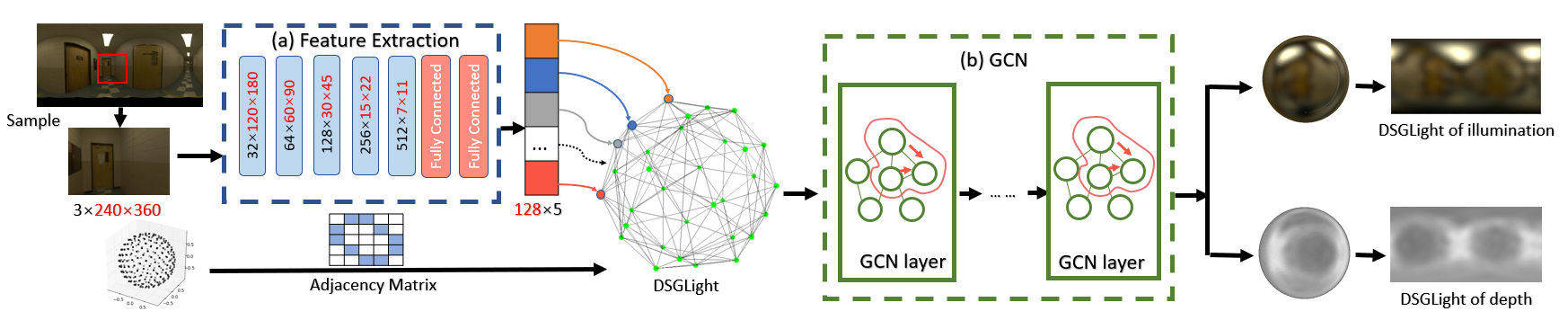}
\caption{The overall pipeline of our method. Given an input image, we first leverage a backbone CNN as a feature extraction module (a) to learn the feature of nodes in DSGLight. Then, we construct a graph with feature vectors initialized by the feature extraction module. A GCN (b) is employed to exchange feature vectors across adjacent nodes and finally predicts the values (color or depth) of the SG nodes.}
\label{fig:network}
\end{figure*}

\section{Deep Graph Learning for DSGLight}\label{sec4}

Our goal is to recover DSGLight of the environment in which the photograph has been taken. It does not simply assign light intensity and depth value to each visible light source, but must extrapolate large portions of the invisible illumination. Thus, it requires wide context information and must exploit very specific relationships between light sources to guide estimation.
To this end, we exploit a graph convolutional network (GCN) to use prior knowledge on the non-Euclidean nature of graph structure of DSGLight and cope with significant amounts of potential relationships between nodes.
Running graph convolutions updates the indoor-specific perceptual features extracted from the input image, which is equivalent as driving light sources' intensities towards the correct light intensities in the RGB color space. 

\subsection{Network architecture}

Our framework consists of two parts: a feature extraction module and a graph convolutional network (GCN). Fig.~\ref{fig:network} shows the architecture of our network.

In the feature extraction module, we can use any CNN based backbone to extract the features of an input image. Currently, we adopt the VGG-19~\cite{DBLP:journals/corr/SimonyanZ14a} as the backbone and generate a $512\times 7\times 11$ feature matrix given an image with a resolution of $240\times 360$. This feature matrix is further connected to GCN with two fully-connected layers, which transform the feature map to the features of each node in DSGLight. 

In our framework, a graph convolutional layer takes a graph as input and outputs a graph with a different feature matrix.
It can be written as the following non-linear function:
\begin{equation}\label{equ:GCN}
\mathbf{H}^{l+1} = \sigma\left( \mathbf{E}\mathbf{H}^{l}\mathbf{W}^{l}\right) 
\end{equation}
where $\mathbf{H}^{l}$ is the input of the $l^{th}$ layer, $\mathbf{W}^l$ is a trainable weight matrix, $\mathbf{E}$ is the adjacency matrix describing the graph structure in the matrix form, and $\sigma$ is a non-linear activation function (ReLU in our experiments).
Stacking graph convolutional networks can combine graph node features and graph topological structural information to make final predictions.

GCN layer takes the node representations $\mathbf{H}^{l}$ from previous layer as inputs and outputs new node representations. Thus we construct the graph $\mathcal{G}=\{\mathcal{V}, \mathcal{F}, \mathbf{E}\}$ as the input of GCN. GCNs are vertex-based and perform convolution on vertices with edges denoting connections. $\mathcal{V}$ is the set of nodes corresponding to vertices in DSGLight. $\mathcal{F}$ is the features of vertices learned from CNN. $\mathbf{E}$ is the adjacency matrix in Eq. (\ref{equ:GCN}). To get the adjacency matrix, we use k-Nearest Neighbors (k-NN) to construct the directed dynamic edges between nodes.

Our GCN module comprises 4 graph convolutional layers. Li et al.~\shortcite{DBLP:conf/iccv/Li0TG19} stated that stacking more layers into a GCN would easily lead to the vanishing gradient problem and over-smoothing. Our experiments show that four GCN layers are sufficient for multi-scale feature extraction and are stable for training. 

\subsection{Loss functions}
In the design of loss functions, we handle color and depth separately. For depth, we only use L2 loss.
For color, we design a hybrid loss that consists of a reconstruction loss $\mathcal{L}_{R}$, a perceptual loss $\mathcal{L}_{VGG}$ and a weighted L2 loss $\mathcal{L}_{W}$ for the SG parameters, \emph{i.e.}, 
\begin{equation}
\mathcal{L}=\mathcal{L}_{W}+\alpha \mathcal{L}_{R} + \beta \mathcal{L}_{VGG}\label{equ:loss}
\end{equation}
where $\alpha$ and $\beta$ are the weights to balance different losses. In our training, we set $\alpha$ and $\beta$ to 0.2 and 0.1, respectively. 

Reconstruction loss penalizes the per-pixel intensity distance between the ground truth and the predicted environment map reconstructed with Eq. \eqref{hdr represent}. The perceptual loss aims to enhance the perceptual similarity in the predicted illumination. We adopt a VGG-19 architecture which is pre-trained in the image classification task.  Let $\Phi$ be the output of the last pooling layer in VGG-19, and $N_{\Phi}$ represents the total number of pixels in this feature space. The perceptual loss is defined as:
\begin{equation}
\mathcal{L}_{VGG}(p,\hat{p})=\frac{1}{N_{\Phi}}\ell_2(\Phi(p),\Phi(\hat{p}))\label{equ:percep}
\end{equation}
where $p$ and $\hat{p}$ denote the ground truth and the predicted environment map, separately.

L2 loss is used mostly in regression problems but it will lose high-frequency information in training. Since the DSGLight has a wide range of values, we reweight SGs in the DSGLight to boost the importance of salient SGs.
Inspired by ACESFilm (Academy Color Encoding System) curve, which distributes the weights to pixels of HDR images empirically and achieves good tone mapping result, we use the following weighted function:
\begin{equation}
f_{w}(x)=\frac{x (2.51 x+0.03)}{x (2.43 x+0.59)+0.14}\label{equ:ACES}.
\end{equation}
The weighted L2 loss for SG parameters is calculated as :
\begin{equation}
\mathcal{L}_{W}(\boldsymbol{A}, \boldsymbol{\hat{A}})=f_{w}(\boldsymbol{A})\ell_2(\boldsymbol{A}, \boldsymbol{\hat{A}})\label{equ:param}
\end{equation}
where $\boldsymbol{A}$ and $\boldsymbol{\hat{A}}$ represent the ground truth and our predicted parameters, respectively.

\subsection{Discussion on the locality of indoor lighting}
Most lights in indoor scenes have a local influence. This means we not only need to predict the distribution of lighting around current point of view, but also have to estimate the relative positions of the lights. 
To handle depth, we augment DSGLight with depth values. In this case, the lobe amplitude parameter of each SG in DSGLight contains four channels (three for color and one for depth). 

The training requires a large number of panoramas and their ground truth depth maps. However, among our two datasets, only Laval dataset has manually annotated per-pixel depth. From these annotated depth maps, we optimize the SGs of depth according to Eq. \eqref{equ:optimization} where $\mathcal{P}$ is replaced by the depth map. 

Some previous methods, such as ~\cite{li2020inverse} and ~\cite{8954392}, also try to handle spatial variations of lighting for indoor scenes. Unlike ours, they estimate indoor lighting for each inserted point independently, yielding a local representation of lighting that varies spatially. Although they produce finer spacial light prediction, their strategy easily incurs temporal flickering when the inserted virtual object moves in the scene. In contrast, our global representation of indoor lighting guarantees temporal consistency for moving objects. 

\subsection{Training details}
Our network is implemented on top of Tensorflow~\cite{tensorflow2015-whitepaper}. We train it using SGD and the Adam solver with the moment parameters $\beta_1=0.9$ and $\beta_2=0.999$. The learning rate is initially set to $0.001$ and halves every 40 epochs. The weights of all layers in our network are initialized with the Xavier uniform. Training examples are fed into our network with a mini-batch size of 5. We train the network for 150 epochs which takes about two days on one NVIDIA V100 GPU.

%% file: section/ch5.tex
\begin{figure*}[tbp]
  \begin{center}
  \renewcommand\tabcolsep{1.0pt}
  \begin{tabular}{ccccccc}
  \small{Input Image} & \small{~\cite{DBLP:journals/corr/GardnerSYSGGL17}} & \small{~\cite{DBLP:conf/iccv/GardnerHSGL19}} & \small{~\cite{DBLP:conf/cvpr/SrinivasanMTBTS20}} & \small{~\cite{zhan2021emlight}} & \small{Ours} & \small{Ground Truth}\\
    &
    \includegraphics[width=0.14\linewidth, trim={0px, 0px, 0px, 0px}, clip]{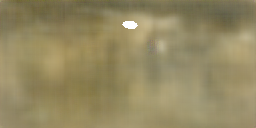}  &
    \includegraphics[width=0.14\linewidth, trim={0px, 0px, 0px, 0px}, clip]{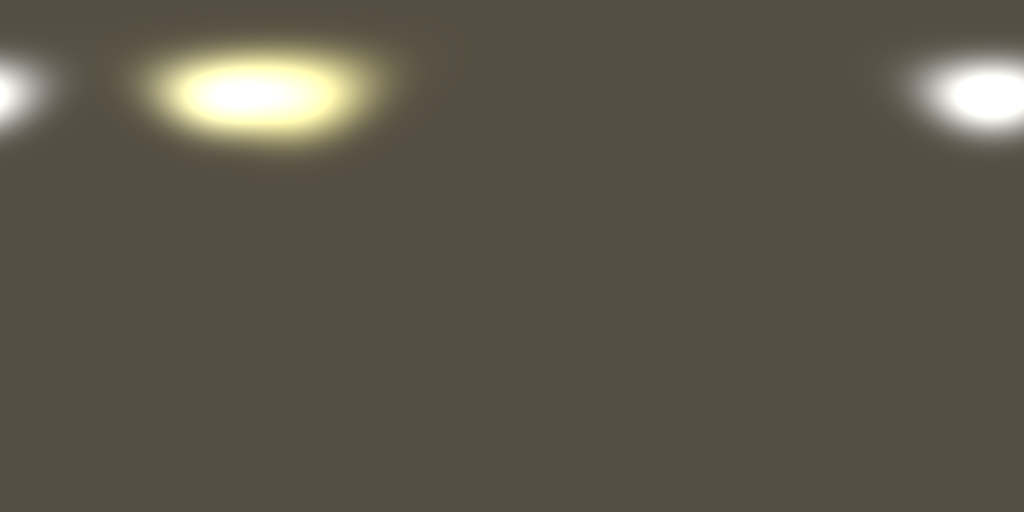} &
    \includegraphics[width=0.14\linewidth, trim={0px, 0px, 0px, 0px}, clip]{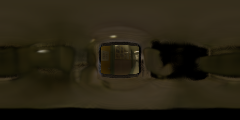}   &
    \includegraphics[width=0.14\linewidth, trim={0px, 0px, 0px, 0px}, clip]{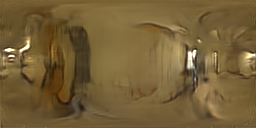}   &
    \includegraphics[width=0.14\linewidth, trim={0px, 0px, 0px, 0px}, clip]{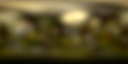} &
    \includegraphics[width=0.14\linewidth, trim={0px, 0px, 0px, 0px}, clip]{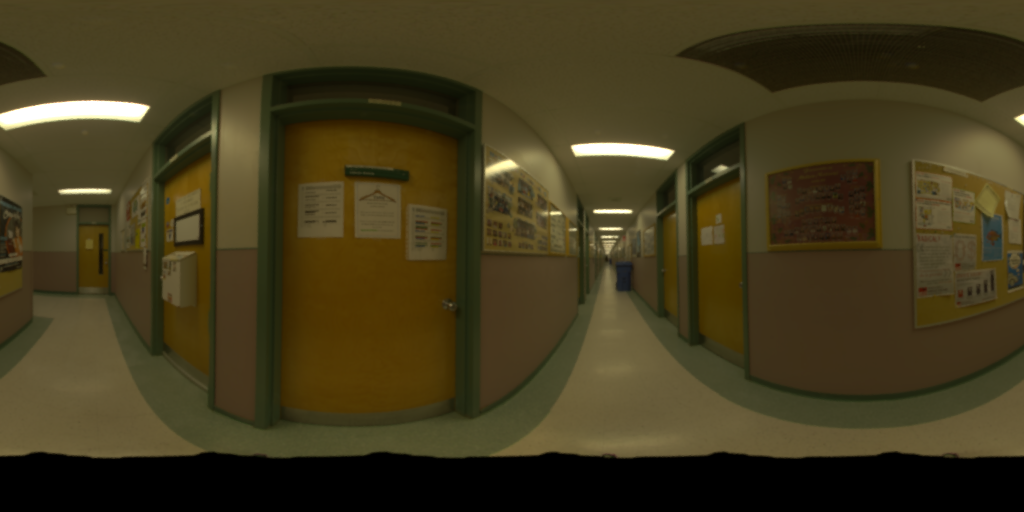}\\
    \includegraphics[width=0.14\linewidth, trim={0px, 0px, 0px, 0px}, clip]{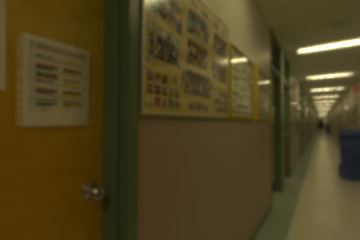}  &
    \includegraphics[width=0.14\linewidth, trim={0px, 0px, 0px, 0px}, clip]{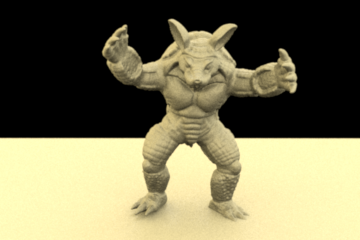}  &
    \includegraphics[width=0.14\linewidth, trim={0px, 0px, 0px, 0px}, clip]{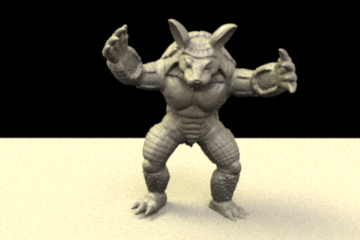} &
    \includegraphics[width=0.14\linewidth, trim={0px, 0px, 0px, 0px}, clip]{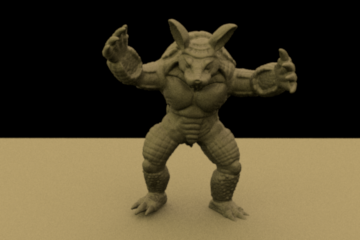}   &
    \includegraphics[width=0.14\linewidth, trim={0px, 0px, 0px, 0px}, clip]{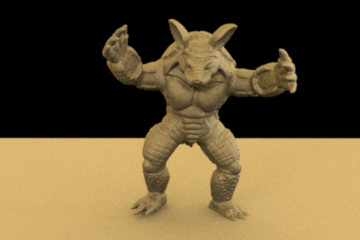}   &
    \includegraphics[width=0.14\linewidth, trim={0px, 0px, 0px, 0px}, clip]{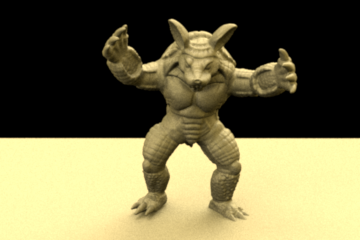} &
    \includegraphics[width=0.14\linewidth, trim={0px, 0px, 0px, 0px}, clip]{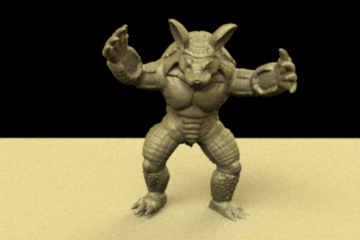}\\
    & \small{18.98~dB} & \small{19.19~dB} & \small{17.09~dB} & \small{21.57~dB} & \small{21.74~dB} &\small{$\infty$~dB}\\
    
       &
    \includegraphics[width=0.14\linewidth, trim={0px, 0px, 0px, 0px}, clip]{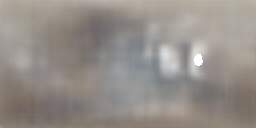}  &
    \includegraphics[width=0.14\linewidth, trim={0px, 0px, 0px, 0px}, clip]{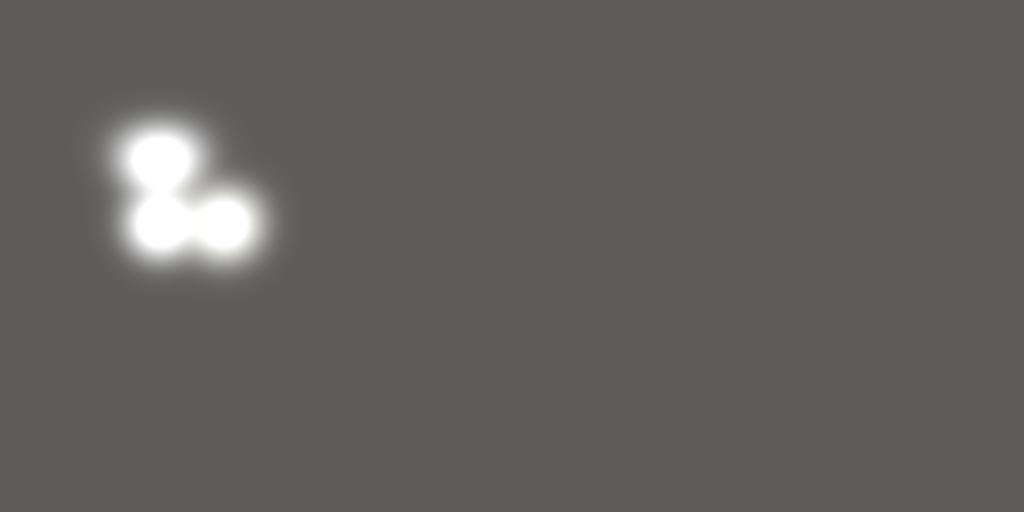} &
    \includegraphics[width=0.14\linewidth, trim={0px, 0px, 0px, 0px}, clip]{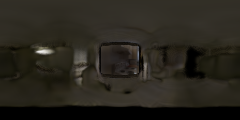}   &
    \includegraphics[width=0.14\linewidth, trim={0px, 0px, 0px, 0px}, clip]{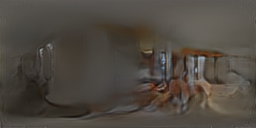}   &
    \includegraphics[width=0.14\linewidth, trim={0px, 0px, 0px, 0px}, clip]{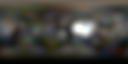} &
    \includegraphics[width=0.14\linewidth, trim={0px, 0px, 0px, 0px}, clip]{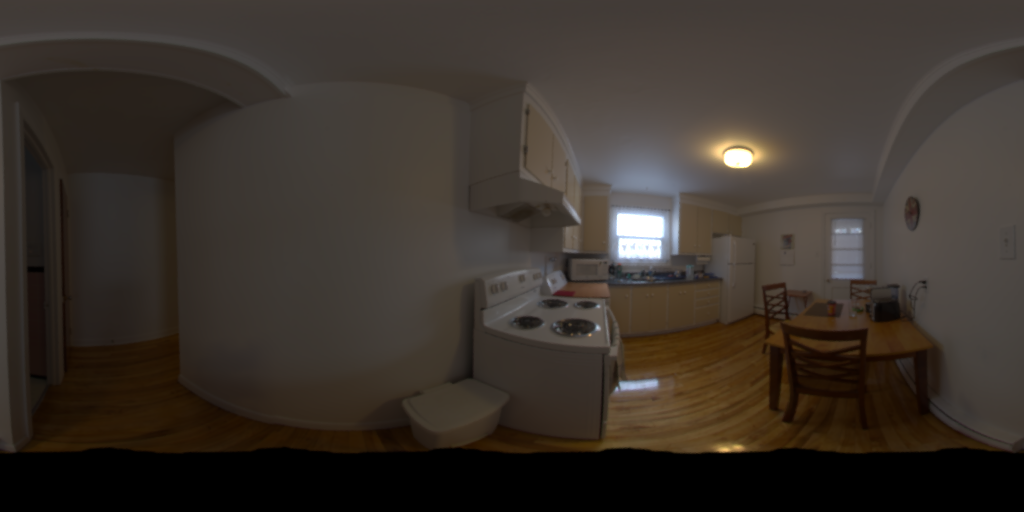}\\
    \includegraphics[width=0.14\linewidth, trim={0px, 0px, 0px, 0px}, clip]{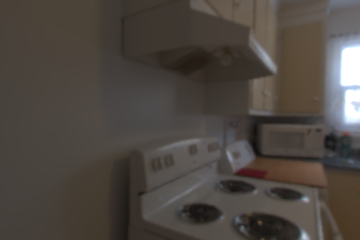}  &
    \includegraphics[width=0.14\linewidth, trim={0px, 0px, 0px, 0px}, clip]{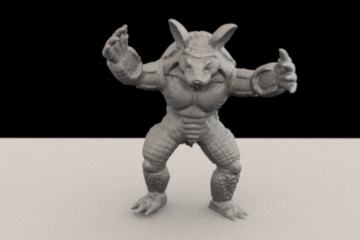}  &
    \includegraphics[width=0.14\linewidth, trim={0px, 0px, 0px, 0px}, clip]{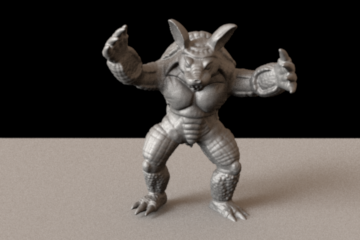} &
    \includegraphics[width=0.14\linewidth, trim={0px, 0px, 0px, 0px}, clip]{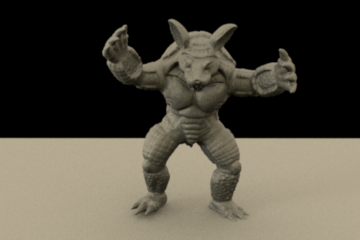}   &
    \includegraphics[width=0.14\linewidth, trim={0px, 0px, 0px, 0px}, clip]{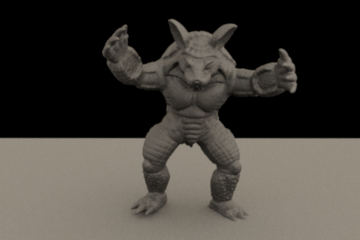}   &
    \includegraphics[width=0.14\linewidth, trim={0px, 0px, 0px, 0px}, clip]{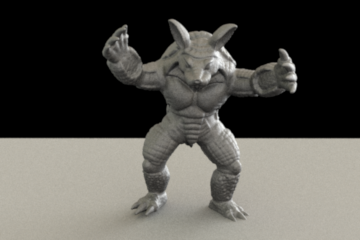} &
    \includegraphics[width=0.14\linewidth, trim={0px, 0px, 0px, 0px}, clip]{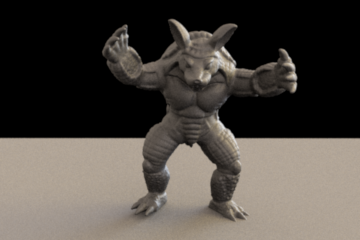}\\
    & \small{18.34~dB} & \small{18.36~dB} & \small{24.51~dB} &\small{23.19~dB} & \small{25.60~dB} &\small{$\infty$~dB}\\

  \end{tabular}
  \end{center}
  \caption{Rendering the Armadillo scene (purely diffuse) with the indoor illumination predicted by different methods. Quantitative metrics in terms of PSNR for the rendered images are displayed below each method.}
  \label{fig:shadow_comparions}
\end{figure*}

\begin{figure*}[tbp]
  \begin{center}
  \renewcommand\tabcolsep{1.0pt}
  \begin{tabular}{ccccccc}
  \small{Input Image} & \small{~\cite{DBLP:journals/corr/GardnerSYSGGL17}} & \small{~\cite{DBLP:conf/iccv/GardnerHSGL19}} & \small{~\cite{DBLP:conf/cvpr/SrinivasanMTBTS20}} &\small{~\cite{zhan2021emlight}} & \small{Ours} & \small{Ground Truth}\\
    &
    \includegraphics[width=0.14\linewidth, trim={0px, 0px, 0px, 0px}, clip]{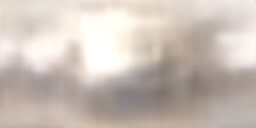}  &
    \includegraphics[width=0.14\linewidth, trim={0px, 0px, 0px, 0px}, clip]{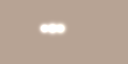} &
    \includegraphics[width=0.14\linewidth, trim={0px, 0px, 0px, 0px}, clip]{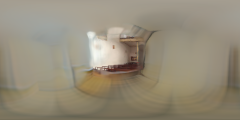}   &
    \includegraphics[width=0.14\linewidth, trim={0px, 0px, 0px, 0px}, clip]{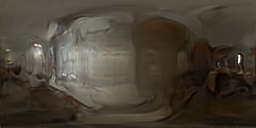}   &
    \includegraphics[width=0.14\linewidth, trim={0px, 0px, 0px, 0px}, clip]{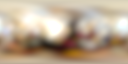} &
    \includegraphics[width=0.14\linewidth, trim={0px, 0px, 0px, 0px}, clip]{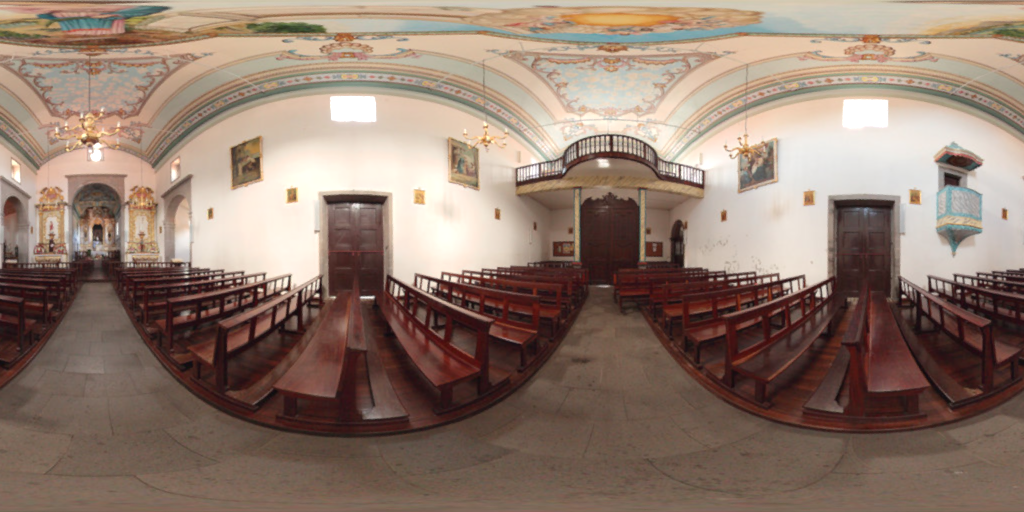}\\
    \includegraphics[width=0.14\linewidth, trim={0px, 0px, 0px, 0px}, clip]{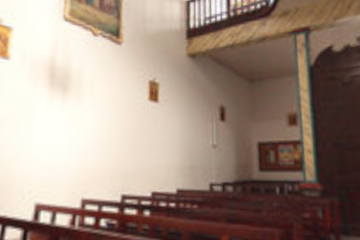}  &
    \includegraphics[width=0.14\linewidth, trim={0px, 0px, 0px, 0px}, clip]{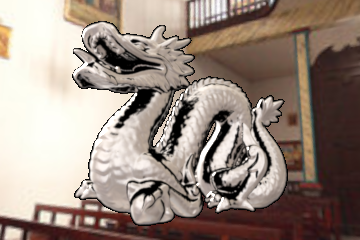}  &
    \includegraphics[width=0.14\linewidth, trim={0px, 0px, 0px, 0px}, clip]{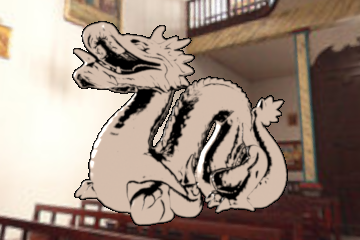} &
    \includegraphics[width=0.14\linewidth, trim={0px, 0px, 0px, 0px}, clip]{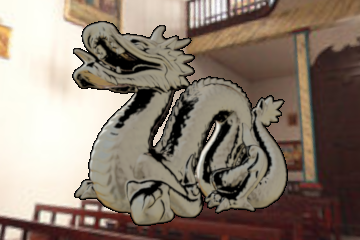}   &
    \includegraphics[width=0.14\linewidth, trim={0px, 0px, 0px, 0px}, clip]{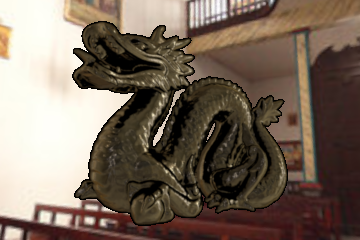}   &
    \includegraphics[width=0.14\linewidth, trim={0px, 0px, 0px, 0px}, clip]{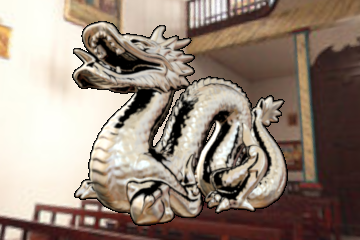} &
    \includegraphics[width=0.14\linewidth, trim={0px, 0px, 0px, 0px}, clip]{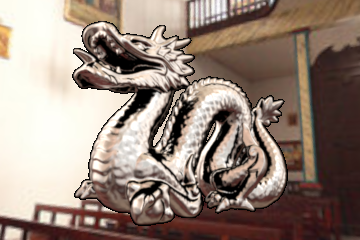}\\
    & \small{19.50~dB} & \small{17.88~dB} & \small{18.33~dB} &\small{12.30~dB} & \small{23.37~dB} &\small{$\infty$~dB}\\
       &
    \includegraphics[width=0.14\linewidth, trim={0px, 0px, 0px, 0px}, clip]{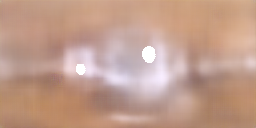}  &
    \includegraphics[width=0.14\linewidth, trim={0px, 0px, 0px, 0px}, clip]{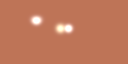} &
    \includegraphics[width=0.14\linewidth, trim={0px, 0px, 0px, 0px}, clip]{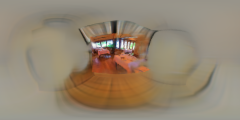}   &
    \includegraphics[width=0.14\linewidth, trim={0px, 0px, 0px, 0px}, clip]{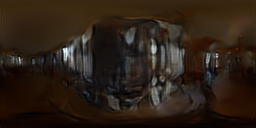}   &
    \includegraphics[width=0.14\linewidth, trim={0px, 0px, 0px, 0px}, clip]{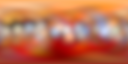} &
    \includegraphics[width=0.14\linewidth, trim={0px, 0px, 0px, 0px}, clip]{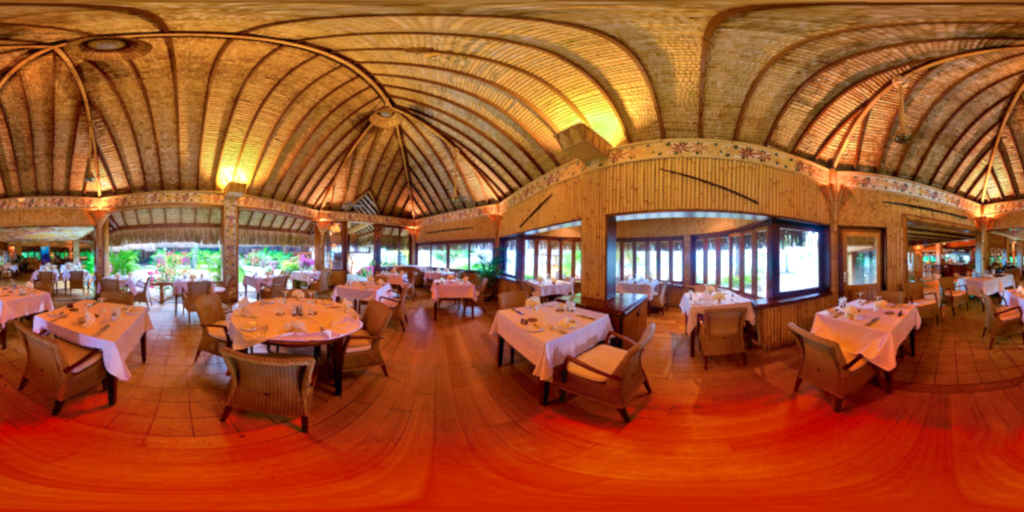}\\
    \includegraphics[width=0.14\linewidth, trim={0px, 0px, 0px, 0px}, clip]{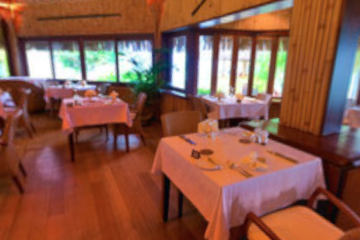}  &
    \includegraphics[width=0.14\linewidth, trim={0px, 0px, 0px, 0px}, clip]{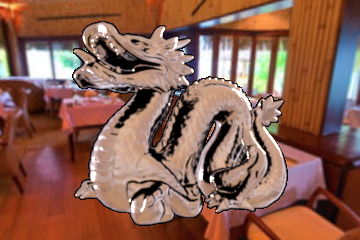}  &
    \includegraphics[width=0.14\linewidth, trim={0px, 0px, 0px, 0px}, clip]{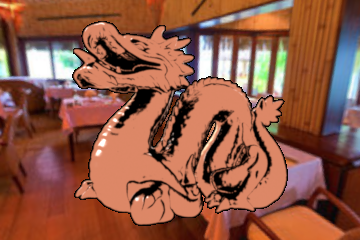} &
    \includegraphics[width=0.14\linewidth, trim={0px, 0px, 0px, 0px}, clip]{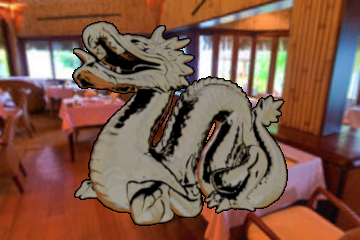}   &
    \includegraphics[width=0.14\linewidth, trim={0px, 0px, 0px, 0px}, clip]{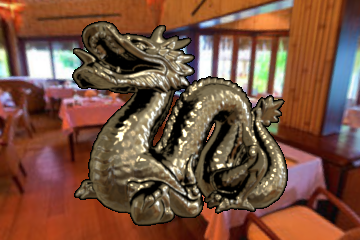}   &
    \includegraphics[width=0.14\linewidth, trim={0px, 0px, 0px, 0px}, clip]{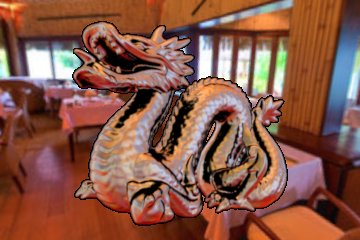} &
    \includegraphics[width=0.14\linewidth, trim={0px, 0px, 0px, 0px}, clip]{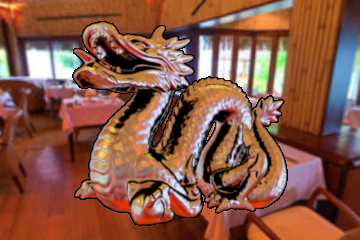}\\
    & \small{19.59~dB} & \small{20.09~dB} & \small{20.28~dB} & \small{18.18~dB} &\small{28.61~dB} &\small{$\infty$~dB}\\
  \end{tabular}
  \end{center}
  
  \caption{Rendering the Dragon scene (specular) with the indoor illumination predicted by different methods. Quantitative metrics in terms of PSNR for the rendered images are displayed below each method. Depth values are not concerned in these cases.}
  \label{fig:envi_comparions}
\end{figure*}

\section{Evaluation}
In this section, we evaluate the performance of our method in indoor illumination prediction and make comparisons with the state-of-the-arts. 
More results and comparisons can be found in the supplemental material.

\subsection{Dataset and metrics}
To generate our dataset, HDR panoramas are subdivided into two parts, one for training and the other for testing. Therefore, the panoramas in the test set will not appear during training. As mentioned in Sec.~\ref{sec:dataset}, we then randomly sample FOV-limited images and generate the ground truth DSGLights from the  panoramas. The SUN360 dataset contains 50,000 images for training and the test set contains 2,000 images. The Laval dataset has 10,000 images for training and 2,000 for testing. 
Visual comparisons are conducted on both predicted environment maps and rendered images with virtual object insertion. PSNR is adopted as the quantitative measure.
\subsection{Comparisons to previous methods}
We predict indoor illumination with different methods and render two scenes in Fig.~\ref{fig:shadow_comparions} and Fig.~\ref{fig:envi_comparions}, respectively. For the methods of ~\cite{DBLP:conf/iccv/GardnerHSGL19} and ~\cite{zhan2021emlight}, we implemented and trained their networks with two datasets separately. Lighthouse~\cite{DBLP:conf/cvpr/SrinivasanMTBTS20} requires a held-out perspective view near the input stereo pair and the depth. Since SUN360 lacks ground truth depth, we only retrain Lighthouse on the Laval dataset. As for evaluation on SUN360, we use their pre-trained model. More training details of Lighthouse are provided in the supplemental material. Rendering is performed with a physically-based Mitsuba renderer~\cite{Mitsuba}.

\textbf{Qualitative evaluation on the Laval dataset:}
In Fig.~\ref{fig:shadow_comparions}, we render the Armadillo scene using the predicted illumination of four recent methods, as well as the ground truth and our method. The renderings of the Armadillo on a plane highlight the accuracy of estimated light directions. Note that our generated shadows and ambient lighting are closer to the ground truth, as compared with other methods. 

Gardner et al.~\shortcite{DBLP:journals/corr/GardnerSYSGGL17} proposed a CNN-based pipeline to predict a RGB panorama and a light mask for the indoor illumination. The predicted panorama is usually very smooth and may have color drift. Therefore, the synthesized images are quite different from those generated by the ground-truth lighting. Gardner et al.~\shortcite{DBLP:conf/iccv/GardnerHSGL19} use a parametric form based on SGs to encode indoor light sources, but the centers of the light sources are allowed to vary wildly, leading to unstable results.
For the method of Srinivasan et al.~\shortcite{DBLP:conf/cvpr/SrinivasanMTBTS20}, it fails to predict the lighting distribution outside the known regions, leading to a large bias on rendering. Zhan et al.~\shortcite{zhan2021emlight} have proved the effectiveness of SGs by generating accurate illumination maps under the guidance of predicted SG parameters. However, they do not support spatially-varying lighting prediction. Most importantly, their CNN networks do not explore the long-range correlation between SGs with a limited receptive field.

\textbf{Qualitative evaluation on the SUN360 dataset:}
In Fig.~\ref{fig:envi_comparions}, we further compare these methods by rendering a specular Dragon model lit by the estimated lighting. Since the model is highly specular, it can clearly reflect the environment around it. As expected, our method captures details in the surroundings (e.g., the white wall and the red floor) quite well.
The predicted panorama is plausible and the details provide better visual coherence. In comparison, Gardner et al.~\shortcite{DBLP:journals/corr/GardnerSYSGGL17} capture the hue of the scene, but the lighting distribution is overly smooth. In contrast, Zhan et al.~\shortcite{zhan2021emlight} hallucinate many details different with the ground truth. For the method of ~\cite{DBLP:conf/iccv/GardnerHSGL19}, it hardly infers and represents lighting details.   

\textbf{Quantitative evaluation}:
We also provide the quantitative metrics in terms of PSNR for each method in Table~\ref{tab:RMSE}.
In this table, we list the average PSNR values for the estimated environment map, the rendered Armadillo scene and the Dragon scene. Obviously, our method achieves higher PSNR values than others under every testing scenario. This further shows that our method achieves state-of-the-art performance on indoor lighting prediction. 

\begin{table}
\caption{\label{tab:RMSE}Quantitative comparisons in terms of PSNR (dB). The best scores are highlighted in bold. -GCN and -$\mathcal{L}_{W}$ represent the network trained without GCN module and weighted L2 loss, respectively.}
\centering
\resizebox{\linewidth}{!}{
\begin{tabular}{l|cccc}
\toprule
Dataset&           Method                 & Map   & Armadillo & Dragon \\ \midrule\midrule
\multirow{7}*{Laval} &~\cite{DBLP:journals/corr/GardnerSYSGGL17}     & 14.45  & 20.60   &  18.35 \\ 
&~\cite{DBLP:conf/iccv/GardnerHSGL19}   &16.49    &  22.21   & 21.14  \\ 
&~\cite{DBLP:conf/cvpr/SrinivasanMTBTS20}     & 17.50 &  21.42  & 23.00  \\
&~\cite{zhan2021emlight}     &  13.80 &   16.63  &  18.76  \\\cline{2-5}
&Ours                  & \textbf{22.54} &  \textbf{26.16}   & \textbf{24.00}  \\
&-GCN                 & 20.90 &  24.83    & 22.11  \\
&-$\mathcal{L}_{W}$                 & 21.87 &  20.30    &  22.61 \\
\midrule\midrule

\multirow{7}*{SUN360} &~\cite{DBLP:journals/corr/GardnerSYSGGL17}       & 11.90 & 23.83  & 20.15 \\ 
&~\cite{DBLP:conf/iccv/GardnerHSGL19}   & 12.36  & 23.58 & 19.04  \\ 
&~\cite{DBLP:conf/cvpr/SrinivasanMTBTS20}        & 13.63 & 21.10     & 21.04  \\
&~\cite{zhan2021emlight}     &  9.02 & 20.17    &  15.97  \\\cline{2-5}
&Ours                  & \textbf{13.75} &   \textbf{24.82}   & \textbf{21.81}  \\
&-GCN                 & 13.16 &  23.39    & 21.21  \\
&-$\mathcal{L}_{W}$                 & 13.70 &  16.85    &  21.54 \\
\bottomrule
\end{tabular}
}
\end{table}

\begin{figure}[tbp]
  \begin{center}
  \renewcommand\tabcolsep{1.0pt}
  \begin{tabular}{cccc}
    \rotatebox{90}{\footnotesize \quad~~ GT} &
    \includegraphics[width=0.31\linewidth, trim={0px, 0px, 0px, 0px}, clip]{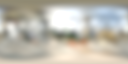} &
    \includegraphics[width=0.31\linewidth, trim={0px, 0px, 0px, 0px}, clip]{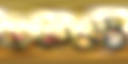}  &
    \includegraphics[width=0.31\linewidth, trim={0px, 0px, 0px, 0px}, clip]{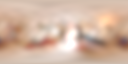}\\
    \rotatebox{90}{\footnotesize \quad Ours} &
    \includegraphics[width=0.31\linewidth, trim={0px, 0px, 0px, 0px}, clip]{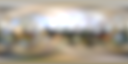} &
    \includegraphics[width=0.31\linewidth, trim={0px, 0px, 0px, 0px}, clip]{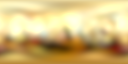}  &
    \includegraphics[width=0.31\linewidth, trim={0px, 0px, 0px, 0px}, clip]{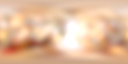}\\
    \rotatebox{90}{\footnotesize \quad -GCN} &
    \includegraphics[width=0.31\linewidth, trim={0px, 0px, 0px, 0px}, clip]{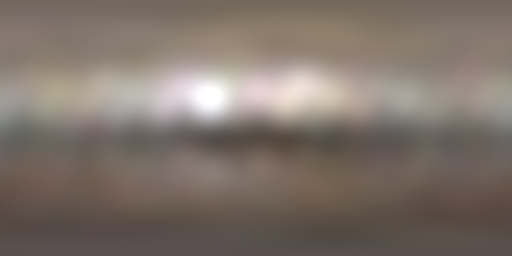} &
    \includegraphics[width=0.31\linewidth, trim={0px, 0px, 0px, 0px}, clip]{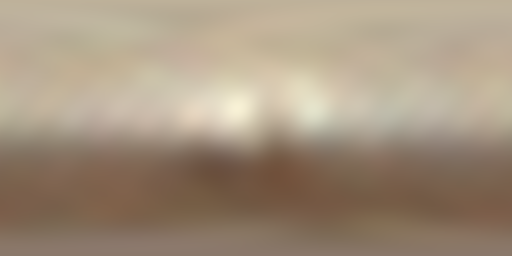}  &
    \includegraphics[width=0.31\linewidth, trim={0px, 0px, 0px, 0px}, clip]{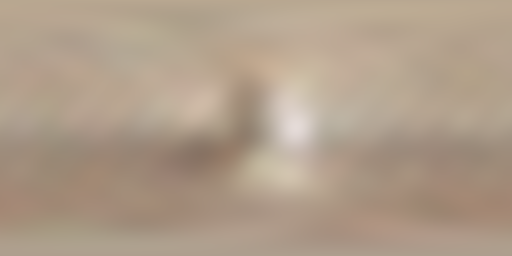}\\
  \end{tabular}
  \end{center}
  \caption{Validating the contribution of GCN to the performance of the presented model.}
  \label{fig:ablation}
\end{figure}

\subsection{Ablation study}

To validate the importance of graph convolutional layers, we remove them from the whole pipeline (-GCN). Now, the features from feature extraction module are fed to a fully connected layer which predicts the parameters directly. 
Fig.~\ref{fig:ablation} reports the performances of these models. 
We observe that environment maps generated by the model without GCN module lack fine details for the scenes. It can not exchange information across neighboring vertices through GCN layers and vertices work independently. Therefore, it is hard to infer illumination from the known region of panoramas. 
Tab.~\ref{tab:RMSE} shows that PSNR drops slightly when our network removes the GCN module. It also shows that weighted L2 loss is beneficial for improving the performance. 
\begin{figure}[tbp]
\centering

\subfloat[~\cite{li2020inverse}]{
    \begin{minipage}[t]{0.3\linewidth}
        \centering
        \includegraphics[width=1.0\linewidth]{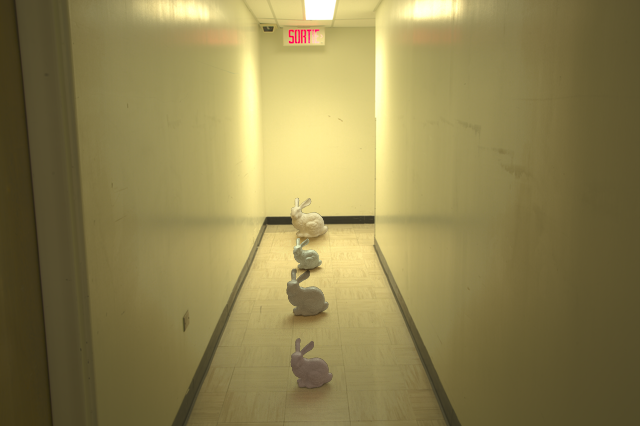}\\
        \vspace{0.02cm}
        \includegraphics[width=1.0\linewidth]{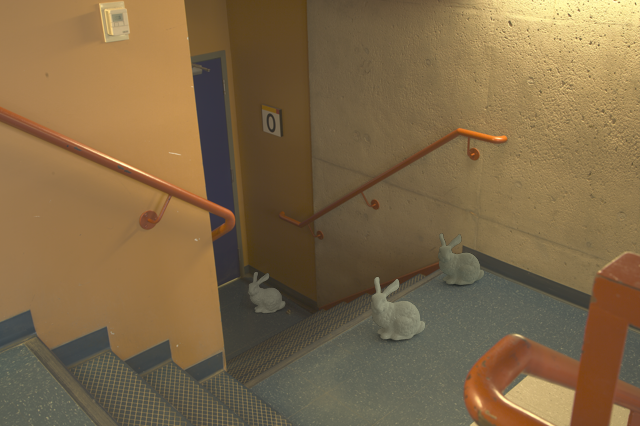}\\
        \vspace{0.02cm}
    \end{minipage}%
}%
\subfloat[DSGLight]{
    \begin{minipage}[t]{0.3\linewidth}
        \centering
        \includegraphics[width=1.0\linewidth]{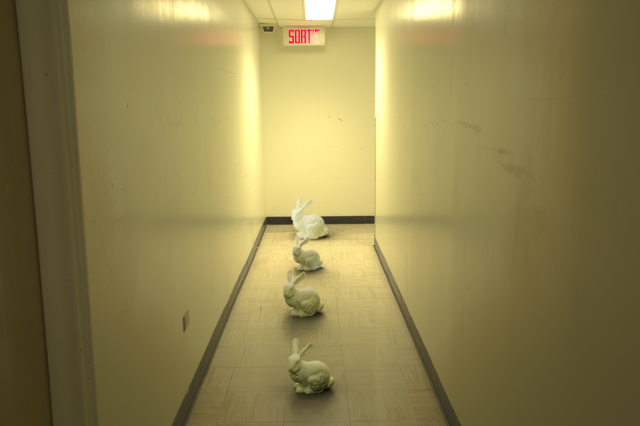}\\
        \vspace{0.02cm}
        \includegraphics[width=1.0\linewidth]{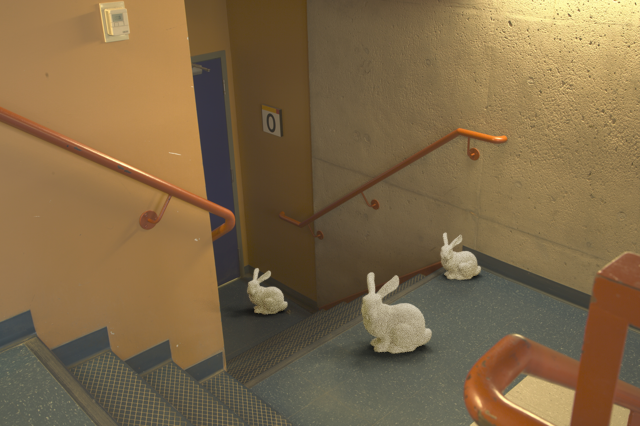}\\
        \vspace{0.02cm}
    \end{minipage}%
}%
\subfloat[Ground truth]{
    \begin{minipage}[t]{0.3\linewidth}
        \centering
        \includegraphics[width=1.0\linewidth]{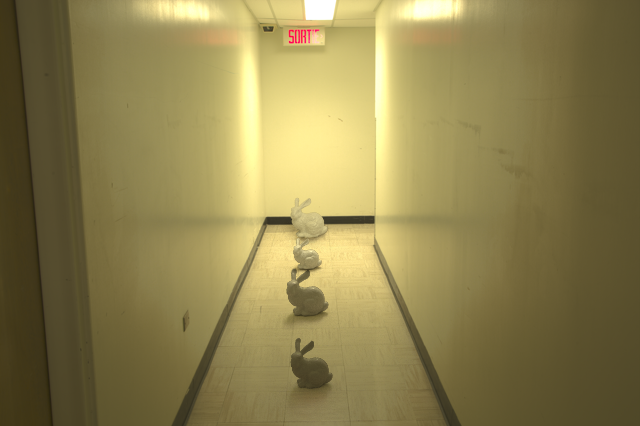}\\
        \vspace{0.02cm}
        \includegraphics[width=1.0\linewidth]{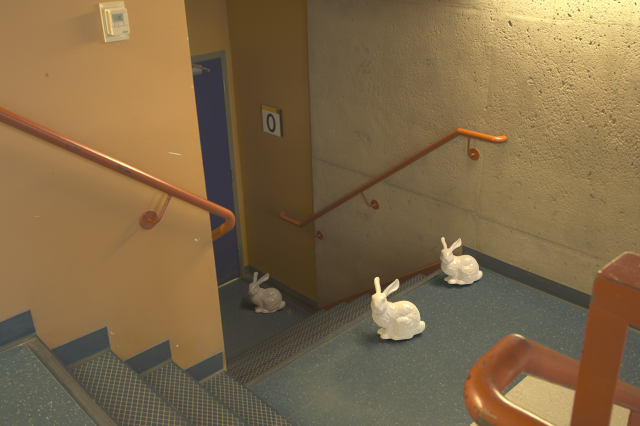}\\
        \vspace{0.02cm}
    \end{minipage}%
}%
\centering
\caption{\label{fig:inserted} Demonstration of spatially-varying lighting on real images of ~\cite{8954392}. 
}
\end{figure}

\subsection{Validation of spatially-varying lighting}
As aforementioned, our method supports spatially-varying lighting when inserting virtual objects into different locations of an indoor scene, due to the availability of the depth information. To demonstrate this, we insert virtual objects at different locations in real images of ~\cite{8954392}. Two examples are provided in Fig.~\ref{fig:inserted}. Here, we compare our method with ~\cite{li2020inverse} that also enables spatially-varying indoor lighting. As seen, our results are closer to the ground truth. More importantly, our method guarantees temporal consistency of lighting when the virtual object moves in the scene, since our DSGLight is unique and global for a given image. However, Li et al.'s method~\cite{li2020inverse} estimates local lighting independently for each inserted position, which may easily cause temporal flickering.

\begin{figure}[tbp]
  \begin{center}
  \includegraphics[width=1.0\linewidth, trim={0px, 0px, 0px, 0px}, clip]{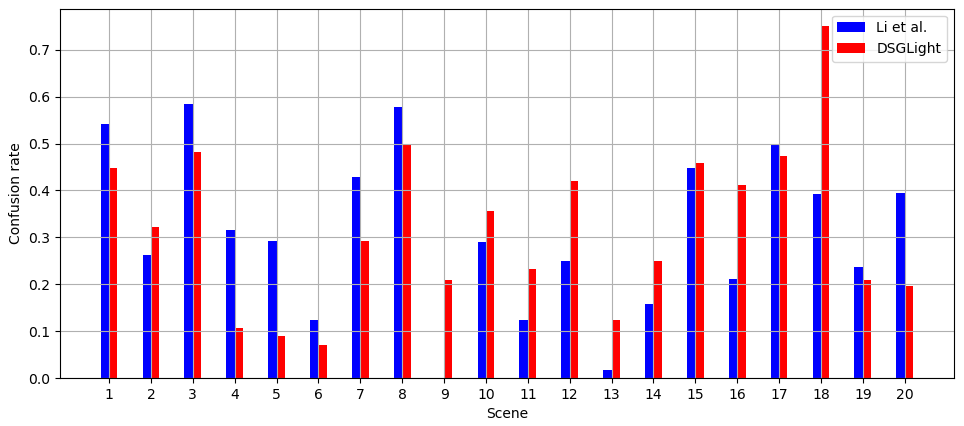} 
  \end{center}
  \caption{Each scene in the user study is shown as a
column, where different colors indicate that users preferred the corresponding method instead
of the ground truth.}
  \label{fig:user study}
\end{figure}

To further show the superiority of our method, we conduct a perceptual user study on 20 scenes of ~\cite{8954392}.
For each scene, we generate a reference composite by relighting a bunny model with its ground truth light probe. We compare these results with objects that were relit with estimated illumination from ~\cite{li2020inverse} and our method. Users are asked to pick which rendering is more realistic between image pairs rendered with ground truth and the estimated lighting.
We gather responses from 138 unique participants and the results are shown in Fig.~\ref{fig:user study}. 
Both of ~\cite{li2020inverse} and our method beat each other in half of the scenarios, which suggests that they are of equal visual quality. However, our method outperforms ~\cite{li2020inverse} slightly on the average confusion, with a 32.00\% vs 30.75\% confusion. 

\subsection{Limitations}
Despite the state-of-the-art performance, our method still has some limitations. First, DSGLight is an approximation to the detailed environment map, lacking sharp features. This prevents specular objects from generating sharp textures reflecting the surrounding. Moreover, the out-of-view depth generated by our method is not very accurate due to the high ill-posedness. This would be solved by multiple input images.

%% file: section/ch6.tex
\section{Conclusion}
In this paper, we have proposed the first graph learning-based strategy for inferring high-quality and spatially-varying indoor illumination from a single image. To represent both dominated local lights and detailed environmental lighting with high fidelity, we encode the panoramic indoor illumination in DSGLight, a depth-augmented lighting model with 128 SGs evenly distributed over a sphere. A GCN is adopted to predict non-Euclidean DSGLight from a 2D image. Through thorough comparisons, we have demonstrated that our method outperforms previous state-of-the-art methods both qualitatively and quantitatively.